\documentclass[sigconf]{acmart}
\usepackage{graphicx} 
\usepackage{booktabs}
\usepackage{array}
\usepackage{enumitem}
\usepackage{multirow}
\usepackage{makecell}
\usepackage{tabularx} 
\usepackage{subfigure}
\usepackage{float}
\usepackage{subcaption}
\usepackage{pgfplots}
\usepackage{amsmath}
\usepackage{amsfonts}
\AtBeginDocument{%
  }

\copyrightyear{2026}
\acmYear{2026}
\setcopyright{cc}
\setcctype{by}
\acmConference[KDD '26]{Proceedings of the 32nd ACM SIGKDD Conference on Knowledge Discovery and Data Mining V.2}{August 09--13, 2026}{Jeju Island, Republic of Korea}
\acmBooktitle{Proceedings of the 32nd ACM SIGKDD Conference on Knowledge Discovery and Data Mining V.2 (KDD '26), August 09--13, 2026, Jeju Island, Republic of Korea}
\acmDOI{10.1145/3770855.3817641}
\acmISBN{979-8-4007-2259-2/2026/08}

\begin{document}

\title{CausalPOI: Spatio-Temporal Graph-Based Causal Modeling for Cold-Start POI Check-in Forecasting}

\author{Zhaoqi Zhang}
\affiliation{%
  \institution{Nanyang Technological University}
  \city{Singapore}
  \country{Singapore}
}

\email{zhaoqi001@e.ntu.edu.sg}

\author{Miao Xie}
\affiliation{%
  \institution{China Agricultural University}
  \city{Beijing}
  \country{China}
}
\email{xiemiao@cau.edu.cn}

\author{Yi Li}
\affiliation{%
  \institution{Nanyang Technological University}
  \city{Singapore}
  \country{Singapore}
}
\email{liyi0067@e.ntu.edu.sg}

\author{Linyou Cai}
\affiliation{%
  \institution{Meituan}
  \city{Beijing}
  \country{China}
}
\email{cailinyou@meituan.com}

\author{Siqiang Luo}
\affiliation{%
  \institution{Nanyang Technological University}
  \city{Singapore}
  \country{Singapore}
}
\email{siqiang.luo@ntu.edu.sg}

\author{Gao Cong}
\affiliation{%
  \institution{Nanyang Technological University}
  \city{Singapore}
  \country{Singapore}
}
\email{gaocong@ntu.edu.sg}

\renewcommand{\shortauthors}{Zhaoqi Zhang et al.}



\begin{abstract}
As urban environments continue to evolve rapidly, accurately modeling the dynamic behaviour of Points of Interest is essential for supporting data-driven urban planning and commercial decision-making. While recent advancements in spatio-temporal graph learning have improved POI forecasting, most methods rely on proximity-based graphs and correlation-driven modeling, which overlook the functional dependencies between POIs and fail to capture the causal effects of urban interventions.
In this paper, we introduce a novel research problem -- cold-start POI check-in forecasting, which aims to predict the future check-in pattern of a newly introduced POI, by modeling its temporal evolution and functional interactions with nearby POIs in a structured urban spatial context. To address these challenges, we propose CausalPOI, a spatio-temporal graph-based causal representation learning framework. CausalPOI leverages Spatio-Temporal Functional Interaction Graph to model semantic and spatial relationships between POIs, and constructs structurally aligned treatment and control graphs to simulate factual and counterfactual scenarios. 
Extensive experiments on real-world SafeGraph datasets demonstrate that CausalPOI significantly outperforms state-of-the-art baselines across the board, validating its effectiveness in spatio-temporal forecasting, semantic interaction modeling, and causal effect estimation, providing a more interpretable and actionable foundation for urban intervention analysis. Source code is available at Github \footnote{\url{https://github.com/ZZQ-NTU/CausalPOI}}.
\end{abstract}



\maketitle

\section{Introduction}
Urban environments are inherently dynamic, characterized by the continuous emergence, evolution, and disappearance of Points of Interest (POIs)~\cite{zhang2024structam} such as restaurants, gyms, libraries, and retail stores. These POIs play a central role in shaping human mobility patterns, influencing economic vitality, and defining the spatial structure of cities. Understanding and forecasting POI-level behaviour is crucial for a wide range of applications, including commercial site selection, transportation planning, and public infrastructure deployment. Traditional forecasting methods have predominantly relied on large-scale spatio-temporal data—such as aggregated mobility flows or region-level check-in statistics—to model urban activity patterns. These approaches~\cite{liu2021urban, yuan2024fine} often utilize coarse-grained spatial features, treating urban space as a collection of homogeneous regions and relying on statistical correlations between them to predict future trends. While such models offer useful insights at the macro level, they struggle to capture the fine-grained dynamics and localized influence of individual POIs—such as temporal bursts in popularity, functional shifts over time, or competitive and complementary effects within dense urban clusters.

Consider a real-world scenario where a new gym is about to open in a busy urban neighbourhood. Before its launch, business operators and digital service platforms seek to answer a key question: How many people are expected to check in at this new facility in the coming weeks? Accurate forecasting of a newly introduced gym’s check-in volume is crucial for assessing membership demand, allocating fitness instructors and equipment, and planning class schedules. Unlike conventional forecasting tasks~\cite{yang2018recurrent, hajisafi2023learning, li2024forecasting} that rely on a POI’s historical data, this setting involves predicting user activity for a cold-start POI—one with no prior behavioural observations. To make reliable predictions, the model must infer potential demand as a result of introducing the new gym into an existing urban ecosystem. This requires modeling not only spatial context and functional relationships with nearby POIs, but also understanding how the new POI might causally affect and be affected by its surrounding environment. For example, the presence of competitive gyms may negatively impact the newly added establishment’s popularity, while complementary businesses like smoothie bars may boost foot traffic. Such counterfactual reasoning is essential to distinguish true effects from spurious correlations, and cannot be fully captured by conventional predictive models, since they primarily focus on correlation rather than causality. 

Recent studies~\cite{yu2017spatio, guo2019attention, wu2020connecting, zhang2026urbanmfm} have increasingly focused on modeling urban activity patterns using spatio-temporal data and graph-based approaches. In particular, Graph Neural Networks (GNNs)~\cite{brody2021attentive} have been widely adopted to capture spatial correlations and relational dependencies among POIs, often by constructing graphs based on geographic proximity. These methods have shown success in forecasting POI activity under settings where historical data is available and the urban topology remains relatively static. Additionally, early efforts in causal representation learning have explored the estimation of individual treatment effects (ITE) under the potential outcomes framework~\cite{liu2020estimating, wang2022causalgnn}, offering tools to reason about interventions from observational data.


Despite these advances, two fundamental challenges remain in modeling POI dynamics for decision-making: (1) \textbf{Functional Interaction Modeling.} Existing graph-based forecasting methods mainly rely on spatial proximity to define relationships, overlooking the semantic functionality and real-world interactions among POIs. This simplification limits the ability to model competition and complementarity, which are essential for accurate predictions and planning.  
(2) \textbf{Causal Estimation in Structured Spaces.} Although some methods attempt to estimate treatment effects, they often overlook the spatial structure of urban environments and rely on unstructured representations. Consequently, they fail to simulate how interventions, such as introducing a new POI, affect surrounding areas through spatial dependencies.

This paper introduces a novel research problem -- how to model the causal effect of a newly introduced POI on urban mobility patterns by explicitly representing its temporal evolution and functional interaction with surrounding POIs, which can be formulated as a cold-start POI check-in forecasting problem. Different from prior region-level flow prediction or standard graph forecasting tasks, our setting focuses on POI-level intervention-aware forecasting under localized spatial interference, where the introduction of a new POI changes the local functional interaction context. To address the aforementioned challenges, we propose CausalPOI, a spatio-temporal graph-based representation learning framework for POI-level forecasting. CausalPOI integrates two key components: the Spatio-Temporal Functional Interaction Graph (ST-FIG) Module, which captures both spatial proximity and semantic functional relations among POIs, and the Causal Inference Module, which estimates the individual treatment effect (ITE) of newly introduced POIs by simulating counterfactual scenarios. In ST-FIG, interaction strengths are learned via contrastive pretraining, allowing the model to uncover latent semantic dependencies beyond physical distance. The causal module constructs structurally aligned treatment and control graphs and estimates ITE through a shared Graph Neural Network (GNN) encoder and temporal decoder. The key contributions of this paper are summarized as follows:

\begin{itemize}[leftmargin=*]
    \item We define a novel research problem: cold-start POI check-in forecasting, which aims to predict the future check-in pattern of a newly introduced POI, by modeling its temporal evolution and functional interactions with nearby POIs in a structured urban spatial context. This cold-start forecasting setting is critical for real-world decision-making, yet remains underexplored in existing literature.
    \item We propose CausalPOI, a spatio-temporal causal representation learning framework tailored to this task. To the best of our knowledge, this is the first framework that systematically incorporates causal estimation at the POI level by simulating both factual and counterfactual outcomes, enabling fine-grained causal reasoning in urban analysis. CausalPOI comprises two key components: (1) Spatio-Temporal Functional Interaction Graph Module, which captures both spatial proximity and semantic functional relationships among POIs, allowing the model to uncover both competitive and complementary relationships that extend beyond mere geographic distance. (2) Causal Inference Module, which estimates the individual treatment effect of newly introduced POIs based on aligned treatment and control graph structures.
    \item We conduct extensive experiments on cold-start POI check-in forecasting using real-world POI and check-in data from the US. The results demonstrate the effectiveness of CausalPOI in predicting post-introduction POI activity and estimating treatment effects. Compared to baselines, our method achieves notable improvements across all metrics. In particular, CausalPOI achieves up to \textbf{57.8\%} RMSE and \textbf{34.3\%} MAE reduction compared to the best-performing baseline on the most challenging region, demonstrating its robustness under cold-start conditions and its effectiveness in capturing causal effects in dynamic urban environments.
\end{itemize}

\section{Related Work}
\subsection{POI Check-in Prediction} 

POI check-in prediction is commonly framed as a downstream task of urban representation learning, where the central goal is to derive informative region-level embeddings. In this line of research, spatial graphs are typically constructed using POIs, trajectories, or road networks, and the resulting embeddings are employed for tasks such as demand forecasting or mobility analysis. For example, Fu et al.~\cite{fu2019efficient} and Zhang et al.~\cite{zhang2019unifying} model spatial autocorrelations and intra-region structures to support region-level check-in forecasting. More recent approaches, including Wu et al.~\cite{wu2022multi} and Zhang et al.~\cite{zhang2023multi}, further enhance regional representations by incorporating heterogeneous spatial signals through graph-based fusion or contrastive learning techniques. Li et al.~\cite{li2020competitive} also investigate POI interactions by analyzing competitive relationships, highlighting the importance of functional dependencies in POI behaviour modeling.

However, these methods primarily focus on regional abstractions, which may overlook the nuanced dynamics of individual POIs. By formulating check-in prediction as a statistical aggregation at the regional level, they often fail to capture the evolving patterns of specific POIs—such as temporal bursts, localized trends, or functional shifts. A few studies attempt to directly model POIs. For instance, Tschernutter et al.~\cite{tschernutter2021latent} predict POI-level check-in volumes by modeling latent interactions among POIs. Nonetheless, these models generally assume static POI behaviour and lack the ability to capture temporal evolution or structural changes in local contexts.


\subsection{Causal Modeling} 
Conventional POI forecasting models predominantly rely on statistical correlations observed in spatio-temporal data, limiting their capacity to uncover causal effects—especially in counterfactual scenarios, such as evaluating the impact of adding or removing a POI. To overcome these limitations, causal representation learning has recently garnered attention. Pioneering works such as TARNet and CFRNet~\cite{shalit2017estimating}, and DragonNet~\cite{shi2019adapting}, leverage the potential outcomes framework to estimate ITE by learning representations that reduce confounding bias. However, these methods are primarily designed for tabular or sequential data and are not well-suited for structured spatial domains.

Meanwhile, graph-based causal inference has emerged as a promising direction for relational data. For example, CausalGNN~\cite{wang2022causalgnn} proposes to learn causal representations on static graphs by integrating intervention modeling with message passing, while HyperSCI~\cite{ma2022learning} extends this line of work to higher-order relational structures. Yet, such methods often assume fixed graph topologies and lack mechanisms to handle dynamic spatial structures, which are common in urban environments. Additionally, they typically neglect the temporal evolution of spatial dependencies and the functional semantics of urban entities like POIs.


\section{Preliminary}
\noindent \textbf{Definition 1} \textit{Point of Interest}: A POI $p$ refers to a point with a geographic position, $p.g = (lat_p, lon_p)$. It can be associated with 
a set of textual tags $p.t = (t_1, t_2,...,t_i)$. 
\\ \textbf{Definition 2} \textit{POI Check-in sequence}: Each POI $p$ has a check-in sequence $p.c=\{c_1, c_2, ...,c_W\}$, and each check-in record is denoted as a tuple $c = (w, n)$ indicating that the POI $p$ was visited $n$ times in the week $\mathrm{w}$.
\\ Examples of POI and check-in sequence are provided in Table \ref{tab: example} in Appendix~\ref{sec: data example}.
\\ \textbf{Problem Statement}: 
We formally define the \textit{Cold-Start POI Check-in Forecasting} problem as follows:  
Given a newly introduced POI $p$ with no historical check-in records, the objective is to 
(1) forecast its future check-in sequence $\hat{p}.c = \{\hat{c}_1, \hat{c}_2, \dots, \hat{c}_\mathrm{W}\}$ over the next $\mathrm{W}$ weeks; and  
(2) estimate the individual treatment effect (ITE) of introducing $p$ by comparing the predicted outcomes under the factual scenario (with $p$ introduced) and the counterfactual scenario (without $p$). During training and evaluation, the ground-truth future check-in sequence $p.c = \{c_1, c_2, ..., c_W\}$ is available as supervision.

\begin{figure*}
    \centering
    \includegraphics[width=0.95\textwidth]{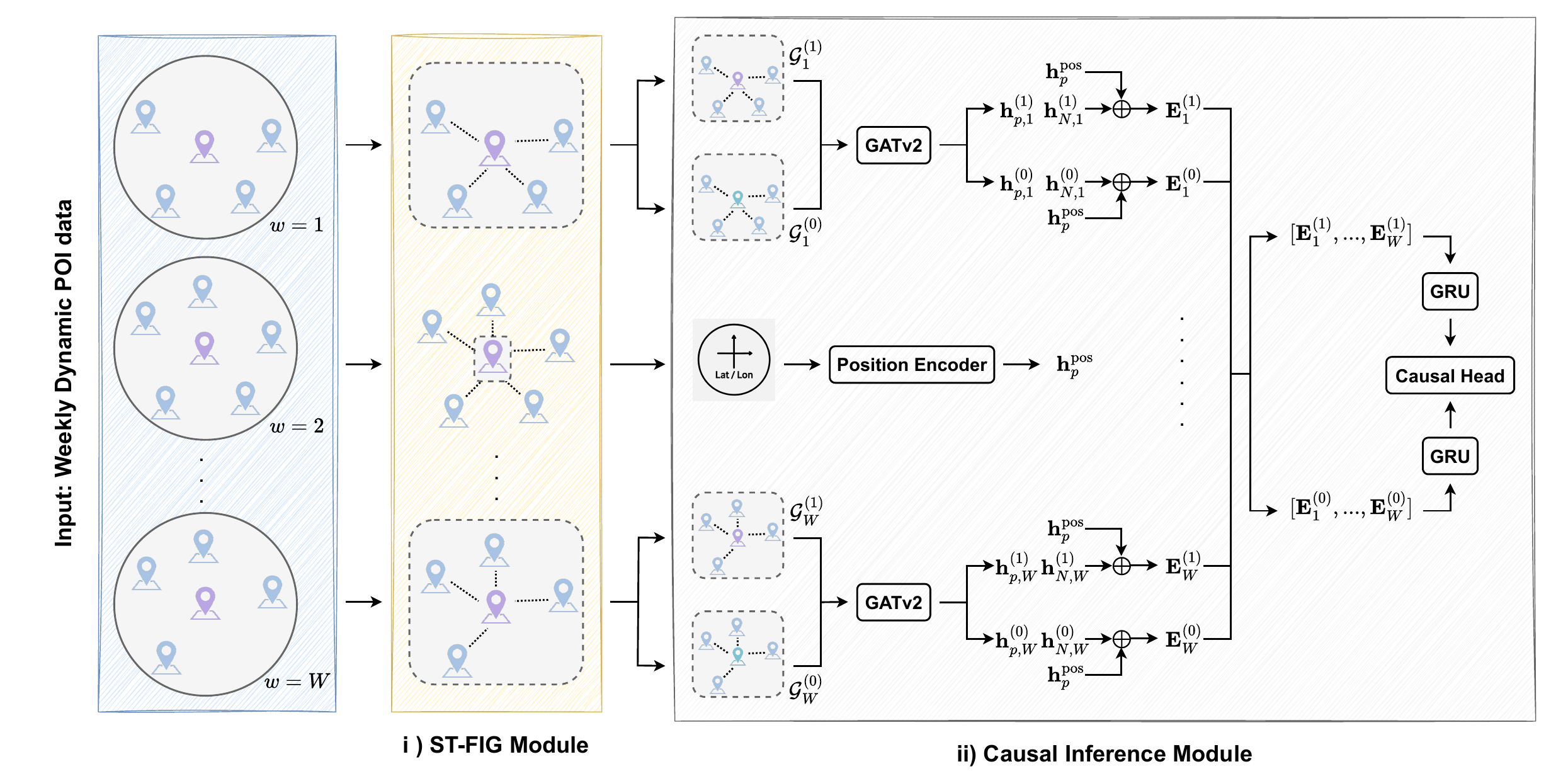}
    \caption{An overview structure of CausalPOI, which consists of two main components: (i) \textbf{ST-FIG Module} where a Spatio-Temporal Functional Interaction Graph is built for each week to capture spatial and semantic relations among POIs; and (ii) \textbf{Causal Inference Module} where treatment and control graphs are encoded using shared Graph and Position Encoders, and their temporal evolution is modeled using a GRU to estimate the causal effect of the newly introduced POI.}
    \label{fig: process}
\end{figure*}

\section{Methodology}
We model the \textit{cold-start POI check-in forecasting} problem as a time series problem, which can better capture the trends and dynamic behaviours of POIs over time. Time series modeling accounts for temporal dependencies, identifying patterns in the evolution of POIs, and making more accurate predictions. To this end, we propose CausalPOI, a spatial-temporal graph-based causality-aware representation learning framework tailored to estimate the Individual Treatment Effect of newly introduced POIs. Unlike traditional approaches that rely on region-level aggregation, CausalPOI focuses on POI-level time series, enabling the model to recognize fine-grained behavioural patterns specific to different time periods. 
It explicitly models the counterfactual scenario in which the POI had not been introduced, allowing the framework to disentangle causal influence from mere correlation. 
Although our implementation builds upon standard graph representation learning components, the key novelty lies in the intervention-aware problem formulation and the causal graph construction for POI-level forecasting. Specifically, we construct paired treatment-control graphs for each newly introduced POI, enabling localized counterfactual reasoning beyond standard static graph prediction. ST-FIG further captures functional relations among neighbouring POIs rather than relying solely on geographic proximity. 
As illustrated in Figure~\ref{fig: process}, this end-to-end design encourages the learning of spatio-temporal representations that are both temporally consistent and causally informative, thus enabling reliable estimation of the evolving impact of new POIs.

\subsection{Spatial-Temporal Functional Interaction Graph Module}
\label{sec: stg}
The neighbour-based check-in volume captures activity driven by the presence and interaction of nearby POIs, particularly through competitive or complementary relationships. For instance, a gym may boost visits to a nearby smoothie bar due to their functional complementarity, while potentially experiencing decreased check-ins from the emergence of a competing gym nearby. This component reflects the spatial and functional interplay among neighbouring POIs and its influence on individual check-in dynamics. To model these intricate interactions, we construct the \textit{Spatial-Temporal Functional Interaction Graph}, which captures the evolving spatial, temporal, and functional dependencies among POIs. In this graph, each POI is represented as a node, and edges are established based on spatial proximity, temporal co-occurrence, and functional relationships such as competition and complementarity. This comprehensive structure enables the model to learn how nearby POIs influence each other over time, supporting more accurate and context-aware check-in prediction.

Given a newly added POI $p$, we collect a neighbourhood set $N = \{n_1, ...,n_i\}$ comprising nearby POIs that may exhibit competitive or complementary effects due to geographic proximity, as determined by the following condition:
\begin{equation}
\small
    dist_{(p, n)} = Haversine(lat_p, lat_n, lon_p, lon_n, r) \leq max_{dist}
\end{equation}
where the $Haversine$ formula calculates the great-circle distance between two points given their latitudes and longitudes, $r$ is the radius of the Earth, and $max_{dist}$ is set to 1000m. 

We then construct a POI functional interaction graph $\mathcal{G}$, which is constructed as a 1-hop subgraph with the following principles:
\begin{itemize}[leftmargin=*]
    \item \textbf{Node Definition}: Each subgraph $\mathcal{G}$ comprises a POI-type target POI and its spatial neighbours, represented as neighbour-type nodes. Each node is initialized with textual embeddings derived from POI textual tags;
    \item \textbf{Edge Definition}: For each neighbour POI $n \in N$, we construct a directed edge from the neighbour POI $n$ to the target POI $p$, enabling bidirectional message passing. Each edge is assigned a weight $w_{(p,n)}$;
    \item \textbf{Temporal Modeling}: For each newly added POI, we construct four temporal subgraphs corresponding to the first $\mathrm{w}$ weeks following its introduction. This design enables the model to capture early-stage interaction patterns and temporal dynamics of spatial influence. The temporal dynamics in ST-FIG are induced by the week-specific neighbourhood composition and activity context. Specifically, only POIs that are active in week $\mathrm{w}$ are included in $\mathcal{G}_\mathrm{w}$, causing the node set and aggregated neighborhood representations to evolve over time.
\end{itemize}
The textual embedding $E_p$ of the POI $p$ is obtained by encoding its textual tags using pre-trained BERT~\cite{devlin2018bert}:
\begin{equation}
E_p = \text{BERT} \ (\ \text{[CLS]} \  t_1, t_2,..., t_i \ \text{[SEP]} \ )
\end{equation}
where $\text{[CLS]}$ represents the entire input sequence, while $\text{[SEP]}$ is used to separate different segments. 

To define the edge weight $\mathrm{w}_{(p,n)}$ between a newly added POI $p$ and its neighbour $n$, we jointly consider their geographic proximity and functional relationship. Specifically, the weight is computed as:
\begin{equation}
    w_{(p, n)} = \alpha_{(p, n)} \cdot \exp\left(-\frac{dist_{(p,n)}^2}{2\sigma^2}\right)
\end{equation}
where $\sigma (=0.5)$ is a bandwidth parameter controlling the spatial decay, $\alpha_{(p, n)}$ quantifies the functional interaction strength between $p$ and $n$. To estimate $\alpha_{(p,n)}$, we pretrain a POI category encoder using contrastive learning to capture task-specific functional semantics. Specifically, we treat functionally competitive POIs as positive pairs, under the assumption that POIs that share similar categories exhibit stronger competition. In contrast, functionally complementary POIs are considered negative pairs, reflecting weaker semantic alignment and potential cooperation. Each POI category is mapped to a textual description and encoded ultilizing BERT~\cite{devlin2018bert}, trained with the information noise contrastive estimation (InfoNCE)~\cite{oord2018representation} loss:
\begin{equation}
    \mathcal{L}_{CL} = -\frac{1}{B} \sum_{i=1}^{B} \log \frac{\exp(\text{sim}(\mathbf{E}_i, \mathbf{E}_i^+) / \mathcal{t})}{\sum_{j=1}^{B} \exp(\text{sim}(\mathbf{E}_i, \mathbf{E}_j^+) / \mathcal{t})}
\end{equation}
where $B$ is the mini-batch size, $\mathcal{t} (= 0.5)$ is a temperature hyperparameter and $\mathbf{E}_i$ and $\mathbf{E}_i^+$ are the embeddings of the $i$-th anchor and its corresponding positive sample, respectively. All embeddings are derived from POI category descriptions using a pretrained BERT encoder. For each anchor, the remaining positive samples $\{\mathbf{E}_j^+ \mid j\ne i\}$ within the batch are treated as negative samples. After training, we define the functional dissimilarity between a newly added POI $p$ and its neighbour $n$ as:
\begin{equation}
    s_{(p,n)} = -\frac{\mathbf{E}_p \cdot \mathbf{E}_n}{|\mathbf{E}_p| \cdot |\mathbf{E}_n|}
\end{equation}
where $\mathbf{E}_p$ and $\mathbf{E}_n$ denote the category embeddings of POI $p$ and POI $n$ obtained from the pretrained encoder, respectively. This negated cosine similarity ensures that functionally complementary POIs—those with low semantic similarity—receive higher similarity scores, while competitive POIs—those with high semantic similarity—are assigned lower scores. We then normalize this value to the $[0, 1]$ range to compute the functional influence weight:
\begin{equation}
    \alpha_{(p, n)} = \frac{1 + s_{(p,n)}}{2}
\end{equation}

\subsection{Causal Inference Module}
To explicitly reason about interventions, we adopt a networked potential outcome framework with partial interference, where the exposure of each target POI depends on the configuration of its spatially adjacent POIs. The binary treatment indicator $t\in\{0,1\}$ denotes whether a POI is newly introduced, while aggregated neighbour influences are modeled as exposure variables summarizing nearby treatments and contexts. Within this framework, conventional predictive models---which focus mainly on correlation---often fail to disentangle causal effects, especially when assessing how a newly added POI affects and is affected by surrounding check-in dynamics.
To bridge this gap, we introduce a dedicated Causal Inference Module, which is designed to estimate the ITE of a newly added POI by explicitly modeling the counterfactual scenario in which the POI had not been introduced. 

\subsubsection{Causal Effect Formulation}
Unlike traditional causal estimation settings, our goal is to forecast the post-introduction check-in dynamics of a new POI rather than infer a population-level treatment effect. We therefore adopt a localized interference assumption, where the influence of an added POI is constrained within its 1-hop spatial neighbourhood, and interference beyond this range is assumed negligible. This allows us to approximate the counterfactual outcome (absence of the POI) by constructing a control graph with preserved topology but masked semantics. Conditional on spatial and semantic covariates captured by the ST-FIG, the model learns to distinguish true causal influence from spurious correlations, yielding identifiable and interpretable representations for forecasting. 
We acknowledge that site selection can be endogenous because unobserved intent-to-open factors may influence both POI introduction and future demand. Our framework partially mitigates this issue by conditioning on rich pre-treatment local context and comparing treatment and control outcomes under aligned spatial topology. Nevertheless, unobserved confounders cannot be fully captured in the observational setting, and thus the estimated causal effects should be interpreted as localized counterfactual estimates rather than fully randomized intervention effects.

Concretely, the model generates two parallel predictions for a newly introduced POI $p$: $\hat{y}_1$ under the treatment condition, where the semantic influence of $p$ is present, and $\hat{y}_0$ under the control condition, where such influence is removed while preserving the same spatial topology. The contrast between these two predictions provides a counterfactual perspective on the functional impact of $p$, without explicitly optimizing individual treatment effects as supervised targets.


To achieve unbiased estimation, the treatment propensity score is defined solely over pre-treatment covariates. We extract a shared pre-treatment embedding $\mathbf{E}_0$ from the ST-FIG encoder to represent the contextual environment of $p$:
\begin{equation}
 \mathbf{E}_0 = \text{LayerNorm} \left( \text{concat}\left[ \mathbf{h}_{p} \ , \mathbf{h}_{\mathcal{N}} \ , \mathbf{h}_p^{\text{pos}} \right] \right)
\end{equation}
where $\mathbf{h}_{p}$ denotes a zero embedding of the target POI, $\mathbf{h}_{\mathcal{N}}$ is the mean-pooled representation of its neighbour POIs and $\mathbf{h}_p^{\text{pos}}$ is the positional encoding of location $p$ (defined later in Section~\ref{sec:feature}).

To estimate the treatment propensity score, we apply a linear classifier over the shared representation obtained from the temporal graph encoder:
\begin{equation}
\hat{p} = \delta \left( \mathcal{W}_p \cdot \mathbf{E}_0  \right)
\end{equation}
where $\mathcal{W}_p$ is a learnable projection and $\delta(\cdot)$ denotes the sigmoid activation.

\subsubsection{Treatment and Control Graph Construction}
To support this estimation, we construct a pair of structurally aligned spatio-temporal graphs for each POI:
\begin{itemize}[leftmargin=*]
    \item \textbf{Treatment Graph ($\mathcal{G}^{(1)}$)}: A 1-hop subgraph centered around the target POI $p$, incorporating its spatial neighbours. Each node is initialized with its semantic representation.
    \item \textbf{Control Graph ($\mathcal{G}^{(0)}$)}: Shares the same topology as $\mathcal{G}^{(1)}$, but replaces the embedding of the target POI with a zero vector, effectively simulating the absence of $p$ without altering the graph structure.
\end{itemize}
We adopt this feature-masking strategy instead of physically removing the target POI node. Although zero-feature masking is not strictly equivalent to physically removing the target POI, it provides a structure-preserving approximation of the counterfactual scenario by suppressing the semantic influence of the target POI while keeping the local topology aligned between treatment and control graphs. In contrast, directly removing the target node would alter the neighbourhood structure, eliminate its incident edges in the 1-hop subgraph, and introduce structural mismatch. As a result, the treatment-control difference could partly reflect topology changes rather than the semantic effect of the newly introduced POI. By preserving the graph topology, our design maintains structural consistency and enables more reliable counterfactual estimation.

To mitigate potential spillover effects while maintaining comparable relational structure, the control graph keeps neighbour embeddings unchanged and preserves spatial connectivity, but removes the functional interaction term $\alpha_{(p, n)}$ in its edge computation:
\begin{equation}
    w^{(0)}_{(p,n)}=\exp\!\left(-\frac{dist^2_{(p,n)}}{2\sigma^2}\right)
\end{equation}
This design preserves the geographic topology while nullifying semantic effects, thus representing a counterfactual scenario where the target POI exerts no functional impact on its neighbours. For numerical stability, all adjacency weights are locally normalized such that $\sum_{n \in N(p)} \tilde{w}(p,n)=1$, and the normalized $\tilde{w}(p,n)$ is used as the attention bias in the GATv2 encoder for both graphs.

\subsubsection{Feature Extraction}
\label{sec:feature}
To capture both spatial and temporal context, we construct a sequence of weekly graphs over $\mathrm{W}$ consecutive weeks. Each weekly graph $\mathcal{G}^{(t)}_w $ represents the 1-hop neighbourhood of POI $p$ at week $\mathrm{w}$ under treatment condition $t \in \{0,1\}$. 
Unlike static spatial graphs, both the node features and neighbourhood sets are time-varying, reflecting changes in active POIs and week-specific check-in dynamics. 
The neighbour set $\mathcal{N}_\mathrm{w}(p)$ is dynamically updated to include only POIs that remain active during week $\mathrm{w}$, allowing local connectivity to evolve over time.

We use GATv2~\cite{brody2021attentive} as the shared graph encoder for each weekly graph. This choice is motivated by the localized 1-hop structure of ST-FIG, where the key challenge is not deep multi-hop propagation, but edge-aware relation modeling between the target POI and its heterogeneous neighbours. Unlike Transformer or Set Attention, which mainly treat neighbouring POIs as unordered tokens and rely on content-based attention, GATv2 naturally supports message passing with edge attributes. Therefore, we incorporate the scalar edge weight as an edge attribute to directly influence the attention computation, serving as a scaling factor that adjusts the importance of neighbour messages according to their spatial-functional affinities:
\begin{equation}
    \mathbf{H}^{(t)}_w = \text{GATv2} \ (\mathcal{G}^{(t)}_w) \in \mathbb{R}^{(N+1) \times 4d}
\end{equation}
where $N$ denotes the number of neighbour nodes, including the target POI and its spatial neighbours, and $d\ (=128)$ is the embedding dimension. From this matrix, we extract the embedding of the central node $\mathbf{h}_{p, \mathrm{w}}^{(t)}$, corresponding to the target POI $p$, and compute the average embedding of its neighbours:
\begin{equation}
    \mathbf{h}_{\mathcal{N},\mathrm{w}}^{(t)} = \frac{1}{|N|} \sum_{n \in N} \mathbf{h}_{n,\mathrm{w}}^{(t)}
\end{equation}
To incorporate geographic information, we encode the latitude and longitude of each POI using a sinusoidal position encoding scheme that integrates multiple frequency components across dimensions, enabling the representation to capture spatial information at varying levels of granularity. Specifically, for each coordinate $p.g \in \{\text{lat}_p, \text{lon}_p\}$
, the $i$-th dimension of its encoding is defined as:
\begin{equation}
\text{PE}^{(i)}(p.g) =
\begin{cases}
\sin\left( \lambda \cdot p_c \cdot 10000^{-\frac{2k}{d}} \right) & \text{if } i = 2k \\[6pt]
\cos\left( \lambda \cdot p_c \cdot 10000^{-\frac{2k}{d}} \right) & \text{if } i = 2k + 1
\end{cases}
\quad \forall\, p_c \in p.g
\end{equation}
where $d\ (=128)$  denotes the dimensionality of the positional encoding, and $\lambda \ (=100)$ is a scaling factor designed to improve the sensitivity to subtle spatial variations among POIs based on their geographic coordinates. The encodings of the latitude and longitude are concatenated to form the final positional embedding $\mathbf{h}_{\text{pos}}$:
\begin{equation}
    \mathbf{h}_p^{\text{pos}} = \text{Concat}\left( \text{PE}(\text{lat}_p),\ \text{PE}(\text{lon}_p) \right) \in \mathbb{R}^{2d}
\end{equation}
These three components are concatenated and passed through a LayerNorm layer to yield a unified representation for week $\mathrm{w}$:
\begin{equation}
\mathbf{E}_\mathrm{w}^{(t)} = \text{LayerNorm} \left( \text{concat}\left[ \mathbf{h}_{p,\mathrm{w}}^{(t)} \ , \mathbf{h}_{\mathcal{N},\mathrm{w}}^{(t)} \ , \mathbf{h}_p^{\text{pos}} \right] \right)
\end{equation}
The vectors is then fed into a GRU to capture temporal dynamics:
\begin{equation}
\mathbf{E}^{(t)} = \text{GRU}([\mathbf{E}_1^{(t)}, \dots, \mathbf{E}_\mathrm{W}^{(t)}])
\end{equation}
This final representation $\mathbf{E}^{(t)}$ is then used to predict counterfactual check-in sequences under both treatment and control conditions via the causal prediction module.

\subsubsection{Causal Prediction}
The temporally aggregated representation $\mathbf{E}^{(t)}$ encodes both the evolving local interaction context and the geographic characteristics of POI $p$ under treatment condition $t$. 
To enable counterfactual outcome modeling, we adopt a treatment-aware prediction head that maps $\mathbf{E}^{(t)}$ to a sequence of weekly check-in predictions. This design allows the model to generate both factual and counterfactual outcome trajectories while maintaining structural alignment between treatment and control representations. Specifically, $\mathbf{E}^{(t)}$ is fed into two outcome decoders corresponding to treatment and control conditions, respectively:
\begin{equation}
\hat{\mathbf{y}}_t = f_t(\mathbf{E}^{(t)}) \in \mathbb{R}^{W},
\quad t \in \{0,1\}
\end{equation}
where $f_1(\cdot)$ and $f_0(\cdot)$ are parameter-shared temporal decoders with treatment-specific output heads.

\subsection{Training Loss}
The model is trained using the following temporally extended loss:
\begin{equation}
\begin{aligned}
    \mathcal{L}_{TC} = \frac{1}{n} \sum_{i=1}^{n} \Bigg[ 
    & \frac{1}{\mathrm{W}} \sum_{w=1}^{\mathrm{W}} \left( \hat{y}^{(i)}_{t,\mathrm{w}} - y^{(i)}_\mathrm{w} \right)^2 \\
    & + \gamma \cdot \text{CrossEntropy} \left( \hat{p}^{(i)}, t^{(i)} \right) 
    \Bigg],
\end{aligned}
\end{equation}
\begin{equation}
\hat{y}_{t,\mathrm{w}}^{(i)} = t^{(i)} \cdot \hat{y}_{1,\mathrm{w}}^{(i)} + (1 - t^{(i)}) \cdot \hat{y}_{0,\mathrm{w}}^{(i)}
\end{equation}
where $\hat{y}_{t,\mathrm{w}}^{(i)}$ is the predicted outcome at week $\mathrm{w}$ under the observed treatment $t^{(i)} \in \{0,1\}$, and $\gamma(=0.2)$ is a hyperparameter that balances the prediction and treatment components.

\section{Experiments}
To evaluate the effectiveness of CausalPOI, we conduct experiments on four real-world datasets and compare it with established baseline methods. We further perform an ablation study to analyze the contribution of each key component to the overall performance gains. In addition, we conduct parameter sensitivity analysis to examine the robustness of CausalPOI under different hyperparameter settings. Finally, we validate the reliability of the estimated causal effects through an uplift sanity check and a placebo test, which help verify whether the learned effects are meaningful.

\subsection{Experimental Setups}
\subsubsection{Datasets}
In our data collection process, we download POI data and user check-in data from SafeGraph \footnote{\url{https://www.safegraph.com}}. We divide the dataset into four regions, Northeast, Midwest, South, and West, following the American official region segmentation. The data covers the period from Sep 2018 to Jan 2020, which corresponds to exactly 74 consecutive weeks. As part of our experimental design, we define newly added POIs as those that emerged within the observation window and persisted for at least 40 weeks. We focus on predicting the weekly check-in volumes for the first four weeks after each new POI’s introduction. The statistics are summarized in Table \ref{tab: statistic}.

\begin{table}[h]
    \centering
    \renewcommand{\arraystretch}{1.0}
    \caption{Statistics of datasets}
    \label{tab: statistic}
    \scalebox{0.9}
    {
    \begin{tabular}{lccc}
    \toprule
    \textbf{Region} & \textbf{Census} & \textbf{POI} & \textbf{Added POI} \\ [0.5ex]
    \midrule
    Northeast & 42,438 & 998,379 & 16,959 \\ 
    Midwest   & 52,894 & 1,209,427 & 19,116 \\ 
    South     & 75,469 & 2,156,091 & 31,894 \\ 
    West      & 46,938 & 1,453,062 & 21,611 \\ 
    \bottomrule
    \end{tabular}
    }
\end{table}

\begin{table*}[h]
\centering
\caption{Estimation and ablation study of cold-start POI check-in forecasting. We evaluate model performance using Root Mean Squared Error (RMSE) and Mean Absolute Error (MAE). The row highlighted in bold represents the performance of CausalPOI, while the rows above it show the baseline models. Based on their underlying approach, we categorize the baselines into six types: time-series GNN-based (G), node-update-based (N), statistical-based (S), LLM-based (L), causal-based (C) and generative synthesis-based (D). The lower part of the table presents the results of the ablation study. To ensure robustness, all metrics are reported as the mean and standard deviation over ten independent runs.}
\label{tab: pd}
\resizebox{0.8\textwidth}{!}{
\begin{tabular}{ccccccccccc}
\toprule
\textbf{Type} & \textbf{Model} & \multicolumn{2}{c}{\textbf{Northeast}} & \multicolumn{2}{c}{\textbf{Midwest}} & \multicolumn{2}{c}{\textbf{South}} & \multicolumn{2}{c}{\textbf{West}} \\
\cmidrule(lr){3-4} \cmidrule(lr){5-6} \cmidrule(lr){7-8} \cmidrule(lr){9-10}
 & 
 & \textbf{RMSE $\downarrow$} & \textbf{MAE $\downarrow$} 
 & \textbf{RMSE $\downarrow$} & \textbf{MAE $\downarrow$} 
 &  \textbf{RMSE $\downarrow$} & \textbf{MAE $\downarrow$} 
 &  \textbf{RMSE $\downarrow$} & \textbf{MAE $\downarrow$} \\
\midrule
G
& DCRNN 
& \makecell{19.86 \\ \scriptsize{(± 0.36)}} 
& \makecell{4.65 \\ \scriptsize{(± 0.26)}} 
& \makecell{22.40 \\ \scriptsize{(± 0.40)}} 
& \makecell{5.18 \\ \scriptsize{(± 0.30)}} 
& \makecell{23.35 \\ \scriptsize{(± 0.38)}} 
& \makecell{5.39 \\ \scriptsize{(± 0.28)}} 
& \makecell{24.52 \\ \scriptsize{(± 0.35)}} 
& \makecell{5.56 \\ \scriptsize{(± 0.27)}} \\
G
& GraphWaveNet 
& \makecell{18.62 \\ \scriptsize{(± 0.32)}} 
& \makecell{4.41 \\ \scriptsize{(± 0.24)}} 
& \makecell{20.87 \\ \scriptsize{(± 0.36)}} 
& \makecell{4.93 \\ \scriptsize{(± 0.27)}} 
& \makecell{21.83 \\ \scriptsize{(± 0.35)}} 
& \makecell{5.16 \\ \scriptsize{(± 0.26)}} 
& \makecell{23.07 \\ \scriptsize{(± 0.33)}} 
& \makecell{5.31 \\ \scriptsize{(± 0.25)}} \\
G
& AGCRN 
& \makecell{17.10 \\ \scriptsize{(± 0.28)}} 
& \makecell{4.08 \\ \scriptsize{(± 0.22)}} 
& \makecell{19.25 \\ \scriptsize{(± 0.31)}} 
& \makecell{4.59 \\ \scriptsize{(± 0.24)}} 
& \makecell{20.14 \\ \scriptsize{(± 0.30)}} 
& \makecell{4.81 \\ \scriptsize{(± 0.23)}} 
& \makecell{21.36 \\ \scriptsize{(± 0.29)}} 
& \makecell{4.97 \\ \scriptsize{(± 0.23)}} \\
G
& LightST
& \makecell{15.72 \\ \scriptsize{(± 0.30)}} 
& \makecell{3.58 \\ \scriptsize{(± 0.18)}} 
& \makecell{17.66 \\ \scriptsize{(± 0.32)}} 
& \makecell{4.02 \\ \scriptsize{(± 0.19)}} 
& \makecell{18.48 \\ \scriptsize{(± 0.38)}} 
& \makecell{4.25 \\ \scriptsize{(± 0.23)}} 
& \makecell{19.61 \\ \scriptsize{(± 0.33)}} 
& \makecell{4.38 \\ \scriptsize{(± 0.26)}} \\
\midrule
D
& TrafficStream
& \makecell{14.96 \\ \scriptsize{(± 0.27)}} 
& \makecell{3.46 \\ \scriptsize{(± 0.17)}} 
& \makecell{16.98 \\ \scriptsize{(± 0.30)}} 
& \makecell{3.90 \\ \scriptsize{(± 0.18)}} 
& \makecell{17.84 \\ \scriptsize{(± 0.33)}} 
& \makecell{4.12 \\ \scriptsize{(± 0.21)}} 
& \makecell{18.97 \\ \scriptsize{(± 0.31)}} 
& \makecell{4.26 \\ \scriptsize{(± 0.20)}} \\
\midrule
S
& SVGP
& \makecell{15.40 \\ \scriptsize{(± 0.05)}} 
& \makecell{3.25 \\ \scriptsize{(± 0.13)}} 
& \makecell{18.30 \\ \scriptsize{(± 0.04)}} 
& \makecell{4.20 \\ \scriptsize{(± 0.14)}} 
& \makecell{20.80 \\ \scriptsize{(± 0.04)}} 
& \makecell{4.48 \\ \scriptsize{(± 0.16)}} 
& \makecell{22.30 \\ \scriptsize{(± 0.03)}} 
& \makecell{4.79 \\ \scriptsize{(± 0.15)}} \\
S
& LCFM
& \makecell{12.82 \\ \scriptsize{(± 0.03)}} 
& \makecell{2.90 \\ \scriptsize{(± 0.14)}} 
& \makecell{15.88 \\ \scriptsize{(± 0.01)}} 
& \makecell{3.77 \\ \scriptsize{(± 0.15)}} 
& \makecell{17.92 \\ \scriptsize{(± 0.01)}} 
& \makecell{3.98 \\ \scriptsize{(± 0.15)}} 
& \makecell{19.60 \\ \scriptsize{(± 0.01)}} 
& \makecell{4.08 \\ \scriptsize{(± 0.17)}} \\
\midrule
L
& TIME-LLM
& \makecell{9.67 \\ \scriptsize{(± 0.04)}} 
& \makecell{2.08 \\ \scriptsize{(± 0.11)}} 
& \makecell{11.83 \\ \scriptsize{(± 0.03)}} 
& \makecell{2.81 \\ \scriptsize{(± 0.11)}} 
& \makecell{14.06 \\ \scriptsize{(± 0.04)}} 
& \makecell{2.96 \\ \scriptsize{(± 0.11)}} 
& \makecell{15.48 \\ \scriptsize{(± 0.03)}} 
& \makecell{3.12 \\ \scriptsize{(± 0.12)}} \\
L
& TimeCMA
& \makecell{8.88 \\ \scriptsize{(± 0.06)}} 
& \makecell{1.93 \\ \scriptsize{(± 0.10)}} 
& \makecell{10.98 \\ \scriptsize{(± 0.05)}} 
& \makecell{2.62 \\ \scriptsize{(± 0.09)}} 
& \makecell{13.18 \\ \scriptsize{(± 0.07)}} 
& \makecell{2.81 \\ \scriptsize{(± 0.12)}} 
& \makecell{14.62 \\ \scriptsize{(± 0.06)}} 
& \makecell{2.97 \\ \scriptsize{(± 0.11)}} \\
\midrule
C
& GCIM
& \makecell{7.68 \\ \scriptsize{(± 0.10)}} 
& \makecell{2.05 \\ \scriptsize{(± 0.08)}} 
& \makecell{8.34 \\ \scriptsize{(± 0.11)}} 
& \makecell{2.21 \\ \scriptsize{(± 0.09)}} 
& \makecell{9.12 \\ \scriptsize{(± 0.12)}} 
& \makecell{2.29 \\ \scriptsize{(± 0.09)}} 
& \makecell{9.56 \\ \scriptsize{(± 0.10)}} 
& \makecell{2.36 \\ \scriptsize{(± 0.10)}} \\
\midrule
N
& KGDiff
& \makecell{\underline{6.74} \\ \scriptsize{(± 0.09)}} 
& \makecell{\underline{1.93} \\ \scriptsize{(± 0.07)}} 
& \makecell{\underline{7.29} \\ \scriptsize{(± 0.10)}} 
& \makecell{\underline{2.08} \\ \scriptsize{(± 0.07)}} 
& \makecell{\underline{8.01} \\ \scriptsize{(± 0.11)}} 
& \makecell{\underline{2.15} \\ \scriptsize{(± 0.08)}} 
& \makecell{\underline{8.43} \\ \scriptsize{(± 0.09)}} 
& \makecell{\underline{2.21} \\ \scriptsize{(± 0.08)}} \\
\midrule

-
& \textbf{CausalPOI} 
& \makecell{\textbf{5.36} \\ \scriptsize{(± 0.55)}} 
& \makecell{\textbf{1.84} \\ \scriptsize{(± 0.06)}} 
& \makecell{\textbf{5.87} \\ \scriptsize{(± 0.38)}} 
& \makecell{\textbf{2.00} \\ \scriptsize{(± 0.05)}} 
& \makecell{\textbf{5.88} \\ \scriptsize{(± 0.49)}} 
& \makecell{\textbf{1.98} \\ \scriptsize{(± 0.05)}} 
& \makecell{\textbf{6.17} \\ \scriptsize{(± 0.55)}} 
& \makecell{\textbf{1.95} \\ \scriptsize{(± 0.09)}} \\

\midrule
-
& w/o ST-FIG Module
& \makecell{5.89 \\ \scriptsize{(± 0.60)}} 
& \makecell{1.94 \\ \scriptsize{(± 0.07)}} 
& \makecell{6.19 \\ \scriptsize{(± 0.42)}} 
& \makecell{2.05 \\ \scriptsize{(± 0.07)}} 
& \makecell{6.22 \\ \scriptsize{(± 0.50)}} 
& \makecell{2.02 \\ \scriptsize{(± 0.08)}} 
& \makecell{6.31 \\ \scriptsize{(± 0.40)}} 
& \makecell{2.00 \\ \scriptsize{(± 0.06)}} \\
-
& w/o Causal Module 
& \makecell{5.90 \\ \scriptsize{(± 0.44)}} 
& \makecell{1.92 \\ \scriptsize{(± 0.04)}} 
& \makecell{6.13 \\ \scriptsize{(± 0.57)}} 
& \makecell{2.05 \\ \scriptsize{(± 0.07)}} 
& \makecell{6.23 \\ \scriptsize{(± 0.50)}} 
& \makecell{2.03 \\ \scriptsize{(± 0.07)}} 
& \makecell{6.31 \\ \scriptsize{(± 0.28)}} 
& \makecell{1.97 \\ \scriptsize{(± 0.04)}} \\

-
& w/o Propensity Score
& \makecell{5.48 \\ \scriptsize{(± 0.49)}} 
& \makecell{1.88 \\ \scriptsize{(± 0.04)}} 
& \makecell{6.02 \\ \scriptsize{(± 0.31)}} 
& \makecell{2.04 \\ \scriptsize{(± 0.05)}} 
& \makecell{5.98 \\ \scriptsize{(± 0.20)}} 
& \makecell{1.99 \\ \scriptsize{(± 0.04)}} 
& \makecell{6.21 \\ \scriptsize{(± 0.42)}} 
& \makecell{1.96 \\ \scriptsize{(± 0.05)}} \\

\bottomrule
\end{tabular}
}
\end{table*}

\subsubsection{Baselines}
We compare CausalPOI with eleven strong baselines, including four time-series GNN-based models~\cite{li2017diffusion, wu2019graph, bai2020adaptive, zhang2025efficient}, one node-update-based model~\cite{chen2021trafficstream}, two statistical-based models~\cite{naumzik2020mining, tschernutter2021latent}, two large language model (LLM)-based approaches~\cite{jin2023time, liu2025timecma}, one causal-based model~\cite{zhao2023generative} and one generative synthesis-based model~\cite{wang2026knowledge}. Details of the baselines can be found in Appendix~\ref{sec: baseline}.

Our task focuses on the cold-start scenario, where a newly added POI has no historical check-in records available at prediction time. This poses challenges for baseline models that rely on temporal sequences. To enable a practical and model-compatible comparison, we tailor the input configurations according to the intrinsic requirements of each model type. For statistical and LLM-based baselines, which do not rely on temporal behavioural signals, we use only static semantic features. For GNN-based baselines, which require temporal input sequences, we construct a proxy temporal input for each cold-start POI by aggregating check-in records from its spatial neighbours within a predefined radius $max_{dist}$. Specifically, we first compute a neighbourhood-level activity signal by averaging the historical check-ins of neighbouring POIs over several preceding time windows.  Since cold-start POIs typically exhibit very low initial activity, directly assigning the neighbour-averaged signal would substantially overestimate their early-stage check-in volumes. To address this issue, we introduce a cold-start attenuation factor $\omega \in (0,1)$ and scale the neighbourhood-level activity signal accordingly. The attenuation factor $\omega$ is determined using training data only and is defined as a single global scalar, rather than a category-specific or region-specific coefficient. Specifically, we first examine cold-start POIs in the training set and compute their average check-in volume over the initial time windows after introduction, which characterizes typical early-stage cold-start activity. In parallel, we compute the corresponding neighbourhood-level activity signal by averaging the check-in volumes of their spatial neighbours over the same time windows. The attenuation factor $\omega$ is then calibrated to align the magnitude of the neighbourhood-level activity signal with the average early-stage cold-start activity. This training-only calibration prevents information leakage from validation or test data while providing a fair proxy temporal input for baselines that require historical sequences. The scaled sequence is used as the proxy temporal input for the target cold-start POI in GNN-based baselines, while neighbouring nodes use their actual historical check-in sequences or temporally aggregated versions when required by the baseline implementation. A spatial subgraph containing the target POI and its neighbours is then fed into the GNN encoder, and the output representation of the target POI node is used for prediction.

\subsubsection{Experimental Settings}
The model is trained until convergence with a learning rate of 3e-4, a batch size of 32. We utilize the \textit{bert-base-uncased} model with a hidden size of 768 to encode textual information. All the experiments involving deep learning frameworks are executed on a V100-SXM2 GPU.

\subsection{Performance Analysis}
As shown in Table~\ref{tab: pd}, the experimental results demonstrate clear and consistent performance differences among the baseline methods, highlighting their distinct capabilities in handling the cold-start POI check-in forecasting task.

Time-series GNN-based models exhibit the weakest performance across all regions. Although pseudo temporal sequences can be constructed from neighboring POIs under cold-start settings, such approximations introduce substantial noise and fail to capture true behavioral dynamics. Moreover, these models fundamentally rely on historical activity signals and lack mechanisms to incorporate semantic or functional attributes, making it difficult to distinguish functionally different POIs with similar spatial neighborhoods. Statistical-based approaches achieve moderate improvements by incorporating static spatial features. SVGP models spatial influence through distance-based kernels, while LCFM captures latent flows among POIs, highlighting the importance of inter-POI dependencies. However, both methods rely on coarse spatial representations and do not explicitly encode local graph structures, limiting their expressiveness in fine-grained spatial contexts. LLM-based methods substantially outperform the above baselines by leveraging rich textual descriptions to learn discriminative semantic representations, enabling strong generalization under cold-start conditions. Nevertheless, they primarily operate at the semantic level and do not explicitly model spatial proximity or neighborhood interactions.
Among the additional closely related baselines, TrafficStream performs better than time-series GNN models, suggesting that modeling evolving graph structures is helpful under cold-start settings. GCIM and KGDiff further achieve stronger results, indicating the potential benefits of causal spatio-temporal modeling and generative synthesis for unseen nodes. However, these methods still underperform CausalPOI because they are not specifically designed for localized POI-level intervention-aware forecasting or structure-aligned treatment-control comparison. In particular, GCIM mainly focuses on causal representation learning over spatio-temporal structures, while KGDiff emphasizes generating plausible sequences for unseen nodes.

In contrast, CausalPOI consistently achieves the best performance across all regions and metrics. By constructing a local spatial graph centered on the target POI and integrating semantic attributes within a unified causal framework, CausalPOI jointly captures spatial structure and functional semantics, leading to significant performance gains over all competing methods.

\subsection{Ablation Study}
To evaluate the effectiveness of key components in the CausalPOI framework, we conduct ablation experiments by selectively removing the ST-FIG and the Causal Model. 

Table~\ref{tab: pd} reports the ablation results of the proposed CausalPOI framework across four regions. Overall, removing either the ST-FIG module or the causal inference module leads to consistent performance degradation in terms of both RMSE and MAE,  indicating their respective contributions to the model's overall effectiveness. Removing the ST-FIG module results in noticeable performance drops across all regions. Similar degradations are observed in the Midwest, South, and West datasets. This confirms that explicitly modeling functional interactions among neighbouring POIs is crucial. By incorporating semantics-aware interaction strengths into the graph structure, ST-FIG enables the GNN encoder to distinguish functionally relevant neighbours from purely spatially proximate ones, leading to more informative local representations for cold-start forecasting. Removing the causal inference module also consistently degrades performance. This demonstrates that purely predictive modeling is insufficient for estimating the incremental impact of newly introduced POIs under interference. By constructing aligned treatment and control graphs and explicitly modeling counterfactual outcomes, the causal module allows CausalPOI to disentangle genuine treatment effects from correlated neighbourhood dynamics. In addition, we analyze the effect of the propensity loss by setting $\gamma=0$, which corresponds to removing the propensity term from the training objective. As shown in Table~\ref{tab: pd}, removing the propensity loss leads to a small but consistent performance degradation across all four regions. Although the magnitude of degradation is smaller compared to removing the ST-FIG or the causal inference module, the results indicate that the propensity loss provides a stabilizing effect during training. Specifically, it acts as an auxiliary regularizer that encourages balanced representations between treatment and control branches, leading to slightly improved forecasting performance. 


\begin{figure}[t]
\centering
\includegraphics[width=0.43\textwidth]{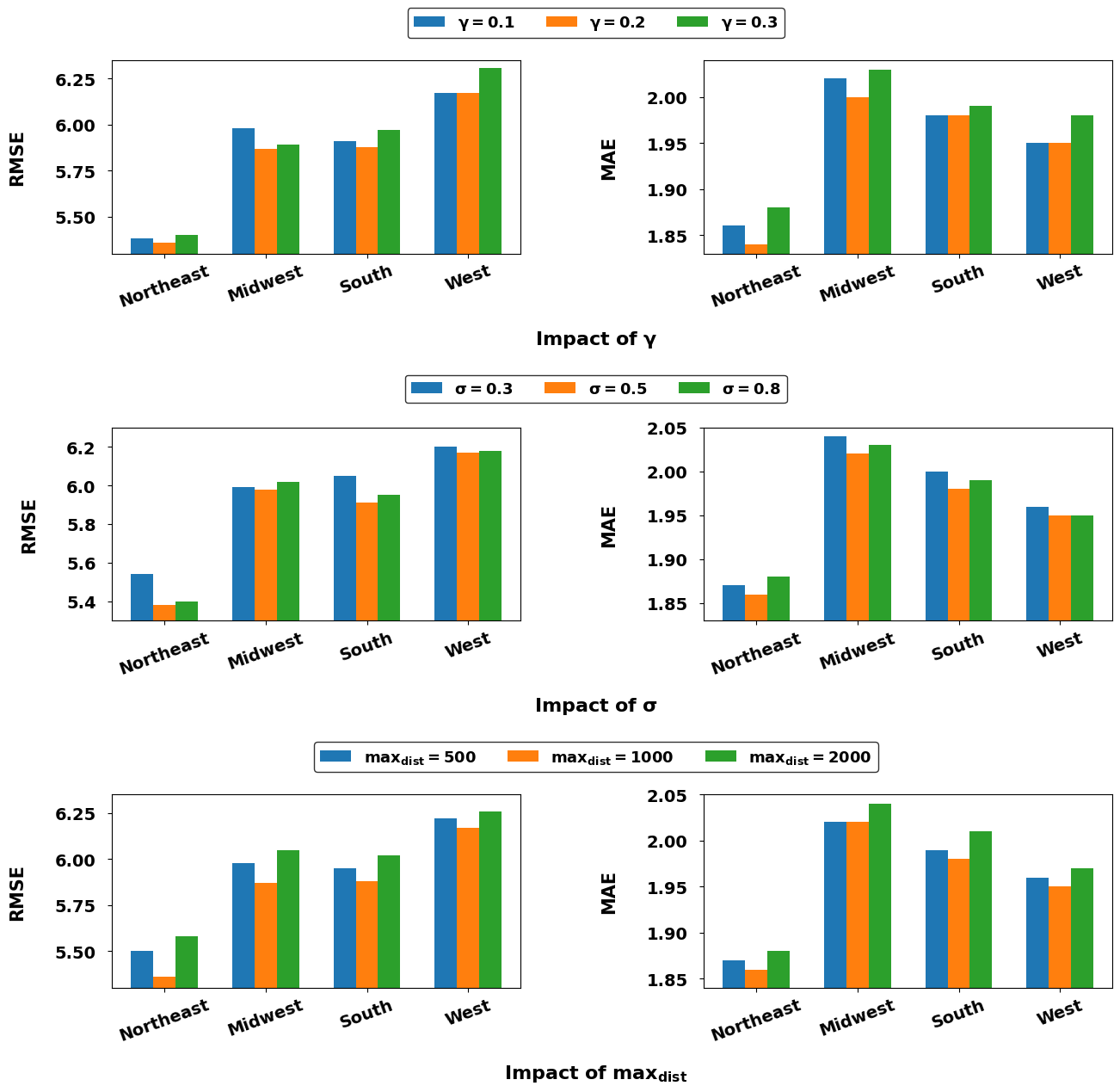}
\caption{Parameter sensitivity analysis of CausalPOIs.}
\label{fig: param}
\end{figure}

\subsection{Parameters Analysis}
We evaluate the sensitivity of CausalPOI to three key hyperparameters: the causal loss weight $\gamma$, the spatial decay bandwidth $\sigma$, and the maximum neighbourhood distance $\max_{dist}$, which respectively control the strength of causal regularization, the attenuation rate of spatial influence in graph construction, and the spatial extent of local neighbourhood graphs. 
As shown in Figure~\ref{fig: param}, the selected parameter configuration achieves the best overall performance across all regions. A too-small value of $\gamma$ weakens the effect of treatment prediction, while an overly large one leads to over-regularization. An overly small $\sigma$ cause spatial influence to decay too rapidly, restricting effective information propagation within local neighbourhoods and reducing the benefit of graph-based modelling, whereas an excessively large $\sigma$ over-smooth spatial influence across distant POIs, diluting locality-aware signals and introducing irrelevant spatial dependencies. Similarly, a too-small $max_{dist}$ restricts the spatial coverage of local graphs, while a too-large value introduces noisy and less relevant neighbours. 

\begin{figure}[t]
\centering
\includegraphics[width=0.43\textwidth]{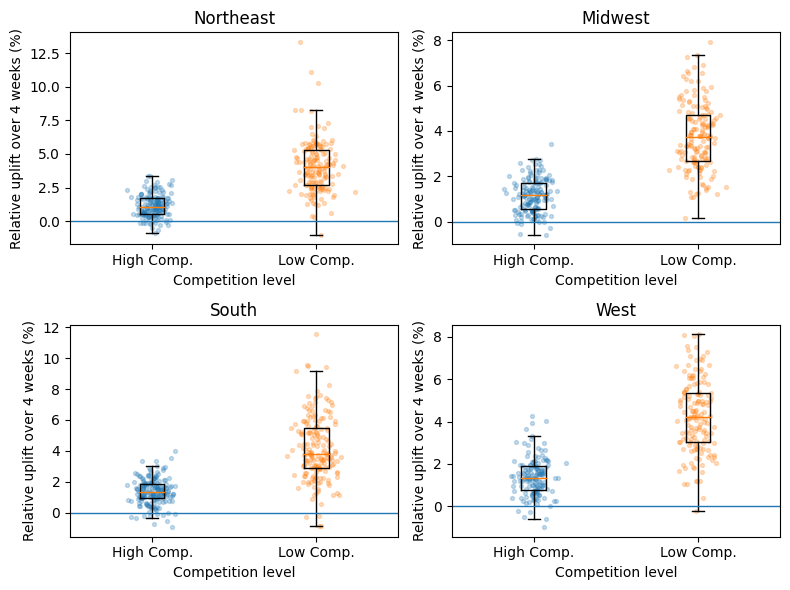}
\caption{Sanity check of estimated uplift of CausalPOI.}
\label{fig: box}
\end{figure}

\subsection{Uplift Sanity Check}
Beyond predictive accuracy, we assess whether the estimated uplift exhibits reasonable behavior under different urban contexts. Since ground-truth causal effects are unavailable, we conduct sanity checks to evaluate the plausibility of the estimated counterfactual outcomes. 
We perform a group-level analysis by comparing the distributions of relative uplift across cold-start POIs under different levels of neighborhood competition. For each POI, we quantify competition by computing the average functional similarity between the POI and its surrounding neighboring POIs, based on semantic embeddings. POIs in the test set of our four datasets are then ranked by this competition score and partitioned into high- and low-competition groups using the top and bottom 20\% quantiles, respectively. As shown in Figure~\ref{fig: box}, POIs in highly competitive environments exhibit consistently smaller relative uplift than those in less competitive settings, which is directionally consistent with urban economic intuition.

\begin{table}[t]
\centering
\caption{Placebo test results of CausalPOI.} 
\label{tab: placebo}
\scalebox{0.8}{
\begin{tabular}{lccc}
\toprule
\textbf{Region} & \textbf{Real Int.} & \textbf{Placebo Int.} & \textbf{Shifted Int.} \\
\midrule
Northeast & 0.228 & 0.013 & 0.171 \\
Midwest   & 0.241 & 0.015 & 0.179 \\
South     & 0.257 & 0.018 & 0.174 \\
West      & 0.249 & 0.016 & 0.173 \\
\bottomrule
\end{tabular}
}
\end{table}

\subsection{Placebo Tests}
To further examine whether the estimated uplift is intervention-specific, we conduct placebo tests under pseudo and shifted intervention settings. As shown in Table~\ref{tab: placebo}, the real intervention setting yields consistently larger uplift values across all regions, ranging from 0.228 to 0.257. In contrast, randomly assigning pseudo-introduction events to non-new POIs leads to near-zero uplift values, ranging from 0.013 to 0.018. This indicates that the model does not assign large treatment effects to arbitrary POIs. When the true introduction week is shifted by approximately four weeks, the estimated uplift also decreases from 0.057 to 0.083 compared with the real intervention setting. These results suggest that the learned uplift is tied to the actual POI introduction event and its timing.

\section{Conclusions}
This paper presents a novel problem -- cold-start POI check-in forecasting, which aims to predict the future check-in pattern of a newly introduced POI, by modeling its temporal evolution and functional interactions with nearby POIs in a structured urban spatial context. To address this challenge, we propose CausalPOI, a spatio-temporal causal representation learning framework that constructs structurally aligned treatment and control graphs to simulate factual and counterfactual outcomes. By introducing the Functional Interaction Graph and leveraging contrastive pretraining, our approach effectively captures both spatial proximity and latent functional semantics among POIs. This design enables precise individual treatment effect estimation and improves predictive performance in POI-level forecasting tasks. Through extensive experiments on real-world datasets, CausalPOI demonstrates strong performance in modeling fine-grained POI dynamics and offers actionable insights for urban planning and intervention analysis.

\begin{acks}
This research is supported by the National Research Foundation, Singapore, under its Frontier CRP Grant (NRF-F-CRP-2024-0005), and under its AI Singapore Programme (AISG Award No: AISG3-RP-2024-034). This research is also supported by NTU SUG-NAP. Any opinions, findings, and conclusions or recommendations expressed in this material are those of the author(s) and do not reflect the views of the National Research Foundation, Singapore. This work is also supported by the China Agricultural University "Young Researcher" Start-up Fund No. QNYJY2024144 and the Visiting Scholar Program of the China Scholarship Council (CSC) No. 202506350123.
\end{acks}

\bibliographystyle{ACM-Reference-Format}
\bibliography{sample-base}

\appendix

\section{Data Example}
\label{sec: data example}
We hereby provide data examples of POI and check-in sequence in Table~\ref{tab: example}.

\begin{table}[h]
\centering
\caption{Examples of POI and POI check-in sequence}
\begin{tabularx}{\linewidth}{@{\extracolsep{\fill}}>{\raggedright\arraybackslash}p{0.1\textwidth}p{0.4\textwidth}}
\toprule
\textbf{Type} & \textbf{Example} \\
\midrule
\textit{POI}  
              & \texttt{id : sg:002a9...bc48e,}\\
              & \texttt{category : Grocery Stores,}\\
              & \texttt{lat : 44.556128,}\\
              & \texttt{lon : -123.066371,}\\
              & \texttt{tags : \{}\\
              & \texttt{name : Jacksons Food Stores,}\\
              & \texttt{street : 33157 Highway 34 SE,}\\
              & \texttt{city : Albany,}\\
              & \texttt{state : OR,}\\
              & \texttt{postcode : 97322}\\
              & \texttt{\}}\\
\midrule
\textit{check-in seq} 
              & \texttt{id : sg:1e57c...2318f,}\\
              & \texttt{category : Grocery Stores,}\\
              & \texttt{lat : 18.465922,}\\
              & \texttt{lon : -66.10359,}\\
              & \texttt{sequence : [}\\
              & \texttt{2018-09-08 to 2018-09-14: 2,}\\
              & \texttt{2018-09-15 to 2018-09-21: 1,}\\
              & \texttt{...} \\
              & \texttt{2020-01-25 to 2020-01-31: 2} \\
              & \texttt{]} \\
\bottomrule
\end{tabularx}
\label{tab: example}
\end{table}

\section{Baselines}

\label{sec: baseline}

\begin{itemize}[leftmargin=*]
    \item \textbf{DCRNN}~\cite{li2017diffusion} models spatiotemporal dependencies using bidirectional diffusion convolution on directed graphs and GRU-based sequence-to-sequence learning, with scheduled sampling to stabilize multi-step forecasting.
    \item \textbf{GraphWaveNet}~\cite{wu2019graph} combines dilated causal convolutions with adaptive graph convolution, learning a self-adjusting adjacency matrix to model latent spatial dependencies and capture long-range temporal patterns.
    \item \textbf{AGCRN}~\cite{bai2020adaptive} introduces node-specific adaptive adjacency matrices and gated recurrent units, enabling flexible modeling of heterogeneous spatial-temporal correlations without fixed graph structures.
    \item \textbf{LightST}~\cite{zhang2025efficient} is a lightweight spatio-temporal graph neural network that improves forecasting efficiency via spatio-temporal distillation while maintaining competitive accuracy.
    \item \textbf{TrafficStream	}~\cite{chen2021trafficstream} focuses on spatio-temporal prediction under node updates, using graph neural networks and continual learning to handle streaming traffic data with dynamically evolving nodes.
    \item \textbf{SVGP}~\cite{naumzik2020mining} uses a sparse Gaussian process to model urban phenomena based on POI distributions, capturing spatial effects through distance-based kernels with interpretable outputs.
    \item \textbf{LCFM}~\cite{tschernutter2021latent} Explains POI check-ins through latent customer flows, including direct visits, competition, and transitions, using a fixed-point formulation and Gaussian processes for interpretable spatial modelling.
    \item \textbf{TIME-LLM}~\cite{jin2023time} reprograms LLMs for time series forecasting via prompts, achieving strong performance without fine-tuning model weights.
    \item \textbf{TimeCMA}~\cite{liu2025timecma} is a recent LLM-based time series forecasting method that enhances prediction by modeling cross-variable temporal dependencies through contextual representations.
    \item \textbf{GCIM}~\cite{zhao2023generative} belongs to causal spatio-temporal intervention modeling, aiming to learn interpretable spatio-temporal representations through generative causal mechanisms.
    \item \textbf{KGDiff}~\cite{wang2026knowledge} is a generative synthesis-based method that leverages knowledge graphs and diffusion modeling to capture heterogeneity and generate spatio-temporal sequences.
\end{itemize}

\end{document}